%% file: camera-ready.tex
\documentclass[10pt,twocolumn,letterpaper]{article}

\usepackage{cvpr}              

\usepackage{multirow}
\input{preamble}

\definecolor{cvprblue}{rgb}{0.21,0.49,0.74}
\usepackage[pagebackref,breaklinks,colorlinks,allcolors=cvprblue]{hyperref}
\usepackage{pifont}
\usepackage{tabularx}
\usepackage{colortbl}
\newcolumntype{Y}{>{\centering\arraybackslash}X}
\usepackage{comment}
\usepackage[accsupp]{axessibility}  

\def\paperID{15341} 
\def\confName{CVPR}
\def\confYear{2026}

\title{PanDA: Unsupervised Domain Adaptation for Multimodal 3D Panoptic Segmentation in Autonomous Driving}

\author{
Yining Pan$^{1,2}$ \quad
Shijie Li$^{2}$ \quad
Yuchen Wu$^{1}$ \quad
Xulei Yang$^{2,*}$ \quad
    Na Zhao$^{1,*}$ \\
$^1$Singapore University of Technology and Design \\
$^2$Institute for Infocomm Research (I2R), A*STAR, Singapore \\
{\tt\small \{yining\_pan,yuchen\_wu\}@mymail.sutd.edu.sg,}\\
{\tt\small  \{li\_shijie,yang\_xulei\}@a-star.edu.sg, na\_zhao@sutd.edu.sg}
}

\begin{document}
\maketitle
{
\renewcommand\thefootnote{}
\footnotetext{* Corresponding authors.}
}
\input{sec/0_abstract}
\input{sec/1_intro}
\input{sec/2_related}
\input{sec_li/3_method}
\input{sec/4-experiment}

\section*{Acknowledgments}
This work was supported in part by the Ministry of Education, Singapore, under its MOE Academic Research Fund Tier 2 (MOE-T2EP20124-0013), and the Agency for Science, Technology and Research (A*STAR) under its MTC Programmatic Funds (Grant No. M23L7b0021). 
{
    \small
    \bibliographystyle{ieeenat_fullname}
    \bibliography{main}
}


\end{document}

%% file: preamble.tex
\usepackage{wrapfig}
\usepackage{graphicx}



%% file: sec/0_abstract.tex
\begin{abstract}
%
This paper presents the first study on Unsupervised Domain Adaptation (UDA) for multimodal 3D panoptic segmentation (mm-3DPS), aiming to improve generalization under domain shifts commonly encountered in real-world autonomous driving. A straightforward solution is to employ a pseudo-labeling strategy, which is widely used in UDA to generate supervision for unlabeled target data, combined with an mm-3DPS backbone. However, existing supervised mm-3DPS methods rely heavily on strong cross-modal complementarity between LiDAR and RGB inputs, making them fragile under domain shifts where one modality degrades (\eg poor lighting or adverse weather). Moreover, conventional pseudo-labeling typically retains only high-confidence regions, leading to fragmented masks and incomplete object supervision, which are issues particularly detrimental to panoptic segmentation. To address these challenges, we propose \textbf{PanDA}, the first UDA framework specifically designed for multimodal 3D panoptic segmentation. To improve robustness against single-sensor degradation, we introduce an asymmetric multimodal augmentation that selectively drops regions to simulate domain shifts and improve robust representation learning. To enhance pseudo-label completeness and reliability, we further develop a dual-expert pseudo-label refinement module that extracts domain-invariant priors from both 2D and 3D modalities. Extensive experiments across diverse domain shifts, spanning time, weather, location, and sensor variations, significantly surpass state-of-the-art UDA baselines for 3D semantic segmentation.

\end{abstract}

%% file: sec/1_intro.tex
\section{Introduction}
\begin{figure}[!h]
    \centering
    \includegraphics[width=\linewidth]{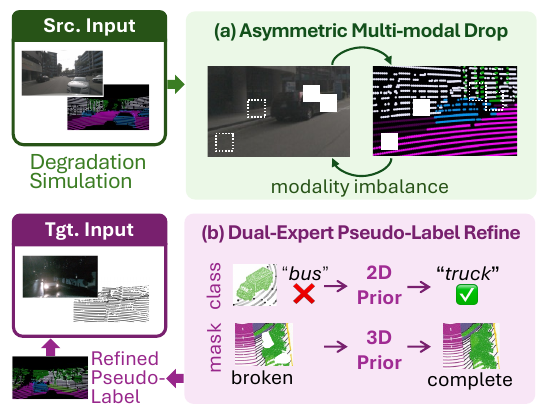}
    \caption{
    Core concepts of our proposed UDA model PanDA. (a) To address modality imbalance arising from challenging conditions in the target domain, we propose \textbf{\textcolor[RGB]{48,79,31}{Asymmetric Multimodal Drop}}, which simulates modality corruption to force cross-modal completion, thereby enhancing overall robustness. (b) To rectify incomplete pseudo-labels, we leverage 2D semantic and 3D geometric priors as a \textbf{\textcolor[RGB]{106,38,103}{Dual-Expert}} to generate dense and reliable supervision, ensuring high-quality panoptic segmentation under domain shifts. 
    }
    \vspace{-0.1in}
    \label{fig:teaser}
\end{figure}

3D panoptic segmentation provides a holistic understanding of scenes by jointly recognizing countable thing instances and amorphous stuff regions, serving as a foundational capability for autonomous driving~\cite{pasco_24cvpr,4dformer_23corl} and robotic interaction~\cite{psg4d_23nips, zu2025collaborative, wangaffordbot}. Recent multimodal approaches combine LiDAR and camera data to achieve more accurate panoptic perception~\cite{lcps_23iccv,pan_2025_ial,p.-fusionnet_24ESA}. However, models often suffer significant performance degradation when deployed in new environments with shifts in location, weather, or time. Since labeling new data is costly and labor-intensive, Unsupervised Domain Adaptation (UDA), which requires no target-domain annotations, offers an appealing solution to improve model generalization across diverse environments.

While UDA has been widely explored in 2D panoptic segmentation~\cite{mansour_LIDAPS_2025,Saha_EDAPS_2023,Ivan_MCPanDA_2024}, 3D object detection~\cite{chen_cmtDA_2024,hu_densityUDA_2023,zhang_PLRefinery_2024,wang2024syn} and 
multimodal 3D semantic segmentation~\cite{wu_2024_unidseg,liang_UniDxMD_2025,jaritz_xMUDA_2020}, it remains unstudied in multimodal 3D panoptic segmentation (mm-3DPS). Directly applying the common pseudo-labeling strategy, which is used to generate supervision for unlabeled target data, is suboptimal for mm-3DPS (as shown in the 4-th row of Table~\ref{tab:main}). This limitation arises from two core challenges.

First, existing mm-3DPS models assume strong cross-modal complementarity between LiDAR and RGB inputs, where both modalities are consistently reliable. This assumption breaks down under real-world conditions such as LiDAR sparsity in rain or poor image quality at night, leading to collapsed cross-modal fusion and degraded scene understanding.
Second, without target-domain labels, pseudo-labeling becomes critical. However, conventional confidence-thresholding~\cite{simons_summit_2023,shin_Mm-tta_2022,xu_vfmseg_2025} often produces fragmented instance masks and blurred boundaries, which severely affect panoptic segmentation quality.

To overcome these issues, we propose \textbf{PanDA}, the first framework addressing multimodal \underline{\textbf{Pan}}optic segmentation under unsupervised \underline{\textbf{D}}omain \underline{\textbf{A}}daptation. Built upon a mean-teacher paradigm~\cite{meanteacher}, PanDA introduces two key components: Asymmetric Multimodal Drop (AMD) and Dual-Expert Pseudo-Label Refinement (DualRefine), as illustrated in Fig.~\ref{fig:teaser}.

AMD simulates real-world modality degradation by selectively masking critical regions in either the LiDAR or image modality within the source domain. Unlike random masking approaches~\cite{zhang_Mx2M_2023,hoyer_MIC_2023,yang_MICDrop_2025}, AMD focuses on panoptic-level critical regions, \ie object interiors and boundaries, encouraging robust, domain-invariant feature learning under artificially imbalanced inputs.

Meanwhile, DualRefine mitigates incomplete and noisy pseudo-labels in the target domain by leveraging both 3D geometric continuity (\ie geometric superpoints) from LiDAR and 2D semantic cues from Vision Foundation Models (VFMs). DualRefine recovers truncated stuff masks using geometric superpoints and corrects thing misclassifications through VFM-informed proposals.

Our main contributions are as follows:
\begin{itemize}
\item We present the first study of UDA for multimodal 3D panoptic segmentation and introduce PanDA, a novel framework enabling robust cross-domain generalization under diverse domain shifts.
\item We propose an asymmetric multimodal drop strategy to simulate realistic modality degradation and enhance domain-invariant representation learning.
\item We develop DualRefine, a pseudo-label refinement strategy that integrates 2D visual and 3D geometric priors to improve instance completeness and label reliability.
\item We conduct extensive experiments across time, weather, location, and sensor shifts, where PanDA achieves significant improvements over state-of-the-art UDA baselines for 3D semantic segmentation.
\end{itemize}

%% file: sec/2_related.tex
\section{Related Work}
\noindent\textbf{3D Panoptic Segmentation.}
LiDAR panoptic segmentation methods can be broadly categorized into top-down, bottom-up, and single-path.
Top-down approaches follow a detection-first paradigm, predicting 3D boxes and then deriving instance masks from the contained points~\cite{effi-lps_21tor,aop_23iv,lidarmultinet_23aaai}. 
Bottom-up approaches instead start with semantic segmentation and generate instances via clustering or grouping~\cite{p-polarnet_21cvpr,dsnet_21cvpr,gps3net_21iccv,scan_22aaai,p-phnet_22cvpr}.
Single-path frameworks treat panoptic segmentation as a unified task, predicting ``thing'' and ``stuff'' masks directly via learnable queries~\cite{maskrange_22arxiv, maskpls_23ral, p3former_25ijcv}.
AdaLPS~\cite{besic_2022_AdaptLPS} represents an early attempt at LiDAR-based UDA for this task, employing handcrafted alignment of ground planes and point densities. However, such case-specific designs generalize poorly to complex domain shifts like weather or geographic variations. 

Motivated by the complementary nature of LiDAR and camera data, recent methods have explored multi-modal fusion. 
LCPS~\cite{lcps_23iccv} and Panoptic-FusionNet~\cite{p.-fusionnet_24ESA} design fine-grained LiDAR-to-image feature fusion, while recent work IAL~\cite{pan_2025_ial} integrates both modalities at the data, feature, and instance levels. Despite these advances, existing methods operate under closed-domain assumptions and lack robustness against modality degradation. 

\begin{figure*}[!h]
    \centering
    \includegraphics[width=1.0\linewidth]{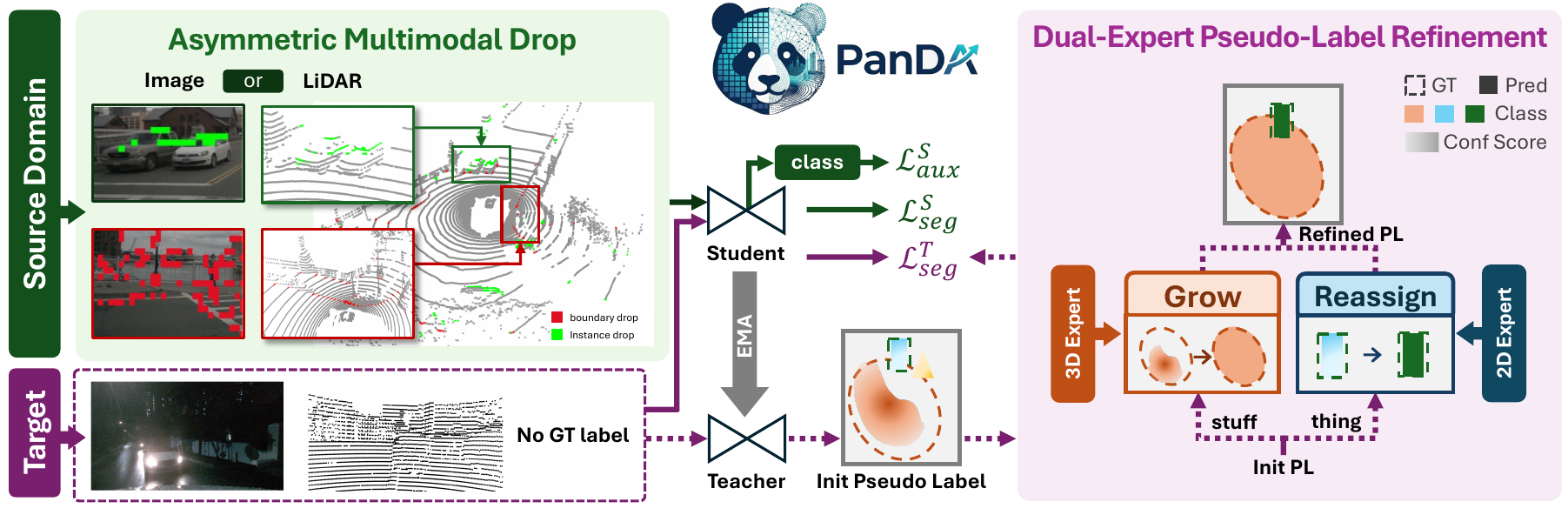}
    \caption{\textbf{The architecture overview of our PanDA framework}. Our model follows a teacher--student paradigm for UDA in multimodal 3D panoptic segmentation. On the labeled source domain (\textbf{\textcolor[RGB]{48,79,31}{green block}}), Asymmetric Multi-modal Drop (AMD) selectively removes instance-interior and boundary regions from either the LiDAR or image modality. 
    This encourages cross-modal recovery and improves robustness to modality imbalance and domain shifts. 
    On the unlabeled target domain (\textbf{\textcolor[RGB]{106,38,103}{purple block}}), Dual-Expert Pseudo-Label Refinement (DualRefine) leverages 3D geometric constraints to refine masks and exploits 2D semantic knowledge to correct category predictions.}
    \label{fig:fw}
\end{figure*}

\noindent\textbf{UDA for Multi-modal 3D Semantic Segmentation.}
In the absence of prior work on UDA for mm-3DPS, we review related studies on domain-adaptive multimodal 3D semantic segmentation, where UDA has been extensively explored.
xMUDA~\cite{jaritz_xMUDA_2020} establishes a foundational paradigm by 
introducing cross-modal distillation to enable knowledge transfer between LiDAR and image modalities, allowing the more reliable modality to guide the weaker one in the target domain. 
Subsequent work extends this along several directions. 
Feature fusion approaches~\cite{wu_cross_2023,cardace_mm2d3d_2023} strengthen interaction between modalities,
while data augmentation techniques, such as style transfer~\cite{man_DualCross_2023}, domain mixing~\cite{chen_comodal_2023}, and rare-object sampling~\cite{cao_MoPA_2024}, enhance diversity and alleviate supervision imbalance. 
Moreover, self-supervised learning strategies, including contrastive learning~\cite{zhang_Exclusive_2022,Xing_contrastive_2023}, adversarial adaptation~\cite{kreutz_LiOnXA_2024}, masked modeling~\cite{zhang_Mx2M_2023} and knowledge distillation~\cite{wu_cross_2023}, have been incorporated to better exploit unlabeled target data. 
Recent methods~\cite{wu_2024_unidseg,Spoeck_2025_vfm,peng_LearningSAM_2024,Joh_ExploringVFM_2025} leverage the strong generalization of VFMs to further reduce domain gaps. 
Beyond using cross-modal loss, UniDxMD~\cite{liang_UniDxMD_2025} learns domain-invariant semantic representations by quantizing heterogeneous multimodal features into a shared latent space. 
Despite their advances, these UDA approaches primarily focus on semantic-level alignment, lacking explicit modeling of instance-level structure. 
This limitation hinders their applicability to panoptic segmentation, which demands both accurate semantic recognition and instance integrity. 


\noindent\textbf{Pseudo-Labeling in Domain Adaptation.} 
Pseudo-labeling is a key mechanism for generating supervision in unlabeled target domains. 
Most UDA methods create pseudo-labels via confidence thresholding~\cite{jaritz_xMUDA_2020}, which often disrupts spatial completeness. 
While CoSMix~\cite{saltori_CoSMix_2022} stabilizes pseudo-labels using a teacher model obtained through temporal ensembling of the student network, and methods like Mm-TTA~\cite{shin_Mm-tta_2022} and SUMMIT~\cite{simons_summit_2023} select labels based on cross-modal agreement. 
Yet these approaches retain only partial high-confidence or high-consistency areas, inevitably fragmenting objects. This issue is particularly harmful for panoptic segmentation, where instance completeness and clear thing–stuff boundaries are crucial. 
Recent methods leverage VFMs to improve pseudo-label quality. 
For example, Xu et al.~\cite{xu_vfmseg_2025} refine class label by voting between model outputs and SEEM~\cite{zou_SEEM_23nips}-inferred labels, while Yang et al.~\cite{yang_samuda_2025} and Zhao et al.~\cite{zhao_SeCoV2_2025} group sparse LiDAR predictions using SAM~\cite{sam_23iccv}. 
In contrast, our method jointly exploits both 2D visual and 3D geometric priors to refine pseudo-labels, improving the reliability of both instance mask and semantic label. 

%% file: sec_li/3_method.tex
\section{Methodology}
\textbf{Task Formulation.}
We address Unsupervised Domain Adaptation (UDA) for multimodal 3D panoptic segmentation (mm-3DPS), where the model is trained on both the labeled source domain $\mathcal{S}$ and the unlabeled target domain $\mathcal{T}$, while evaluated on target domain. 
Each input sample consists of a paired LiDAR point cloud $\mathbf{x}^{3 \mathrm{D}}\in \mathbb{R}^{N \times 3}$  and camera images $\mathbf{x}^{2 \mathrm{D}} \in \mathbb{R}^{H \times W \times 3}$ from both domains. 
Formally, the source and target domains are defined as: $\mathcal{S}=\left\{\left(\mathbf{x}_{\mathcal{S}, i}^{2 \mathrm{D}}, \mathbf{x}_{\mathcal{S}, i}^{3 \mathrm{D}}, \mathbf{y}_{\mathcal{S}, i}\right)\right\}_{i=1}^{|S|}$  and  $\mathcal{T}=\left\{\left(\mathbf{x}_{\mathcal{T}, j}^{2 \mathrm{D}}, \mathbf{x}_{\mathcal{T}, j}^{3 \mathrm{D}}\right)\right\}_{j=1}^{|\mathcal{T}|}$, where $\mathbf{y}_{\mathcal{S},i}$ is the combination of semantic and instance labels in the source domain. 
The objective is to infer semantic and instance labels $\{\mathbf{y}_{\mathcal{T},j}\}_{j=1}^{|\mathcal{T}|}$ for the target domain.

\subsection{Framework Overview}
\label{sec:fw}
In this paper, we propose 
\textbf{PanDA}, the first framework addressing multimodal \underline{\textbf{Pan}}optic segmentation under unsupervised \underline{\textbf{D}}omain \underline{\textbf{A}}daptation, as illustrated in Fig.~\ref{fig:fw}.   
PanDA adopts the mean-teacher~\cite{zhao2020sess, han2024dual, li2024end} paradigm, which consists of a student model and a teacher model, both sharing the same multi-modal 3D panoptic segmentation architecture. Particularly, PanDA uses IAL~\cite{pan_2025_ial} as the segmentation model for its state-of-the-art performance. 
IAL is a transformer-based design that enhances mm-3DPS by leveraging complementary LiDAR and image features in augmentation, feature fusion, and query generation. 
At each training iteration, both source samples $\mathbf{x}_{\mathcal{S}}$ and target samples $\mathbf{x}_{\mathcal{T}}$ are fed to the student model. The source data is supervised by ground-truth semantic and instance labels $\mathbf{y}_{\mathcal{S}}$, while the target data is supervised using pseudo-labels $\hat{\mathbf{y}}_{\mathcal{T}}$ generated and refined from teacher predictions. The teacher model receives only target inputs and is updated by an Exponential Moving Average (EMA) of the student weights. 

The overall training objective consists of three terms: the standard panoptic segmentation loss $\mathcal{L}_{\text{seg}}^{\mathcal{S}}$ and $\mathcal{L}_{\text{seg}}^{\mathcal{T}}$ applied to both source and target domains, a consistency loss $\mathcal{L}_{\text{con}}^{\mathcal{T}}$ for target domain, and an auxiliary semantic loss $\mathcal{L}_{\text{aux}}^{\mathcal{S}}$ for source domain:
\begin{equation}
    \mathcal{L} =\mathcal{L}_{\text{seg}}^{\mathcal{S}}+\mathcal{L} ^{\mathcal{S}}_{\text{aux}} +\mathcal{L}_{\text{seg}}^{\mathcal{T}}+ \mathcal{L}_{\text{con}} ^{\mathcal{T}}.
\end{equation}
The panoptic segmentation loss is the same as the base model IAL. Following the common mean-teacher paradigm~\cite{li_STMono3D_2022,chen_cmtDA_2024}, we introduce a consistency loss to align intermediate representations. Specifically, both teacher and student share the same query initialization for target data, and their updated query features $f_\ell^{(Stu)}$ and $f_\ell^{(Tea)}$ are aligned via $\mathcal{L}_2$ loss at each Transformer decoder layer $\ell$:
\begin{equation}
    \mathcal{L}_{\text{con}}^{\mathcal{T}} = \sum_{\ell} \big\| f_\ell^{(Stu)} - f_\ell^{(Tea)} \big\|_2^2.
\end{equation}
The auxiliary semantic loss $\mathcal{L}_{\text{aux}}^{\mathcal{S}}$ is introduced to enhance masked modeling under our proposed augmentation strategy and will be detailed in Sec.~\ref{sec:Asym}.

PanDA addresses the two fundamental challenges of UDA, \ie domain gap and the absence of target annotations, by jointly exploiting labeled source data and unlabeled target data. 
In Sec.~\ref{sec:Asym}, we introduce an asymmetric multimodal augmentation strategy that simulates modality degradation on the source domain and facilitates the domain-invariant learning on the target domain.  
For the target domain, we first describe in Sec.~\ref{sec:drop} a baseline pseudo-label initiation scheme and analyze its limitations for panoptic segmentation.  
Building upon this, Sec.~\ref{sec:PL} presents our proposed pseudo-label refinement module, which leverages complementary 3D geometric and 2D visual priors to generate complete and reliable pseudo-labels.  

\subsection{Asymmetric Multi-modal Drop}
\label{sec:Asym}
Conventional supervised learning on source data assumes both LiDAR and image modalities are in optimal condition (e.g., clear daytime scenes), making it an "easy" task compared to handling degraded inputs in the target domain (e.g., nighttime or rainy scenes). To bridge this gap and enhance robustness against real-world modality degradation, we introduce \textbf{Asymmetric Multimodal Drop (AMD)}. 
The core idea is to explicitly simulate degradation on one modality in the source domain, and enhance robust cross-modal learning under modality imbalance, and further facilitate to varies realistic target degradations. 
Unlike previous UDA methods that apply random, content-agnostic masking~\cite{zhang_Mx2M_2023,yang_MICDrop_2025,hoyer_MIC_2023}, AMD performs structured dropout based on panoptic-specific properties. Motivated by the distinct characteristics of thing and stuff classes in panoptic perception, we design boundary- and instance-wise dropping strategies.

\noindent\textbf{Boundary Dropout.}  
Accurate boundary perception is essential for distinguishing adjacent objects in panoptic segmentation. 
However, boundary cues often degrade under domain shifts, such as blurred edges in low-light or rainy conditions and unfamiliar geometric structures in new locations. 
 To enhance boundary robustness, we apply boundary dropout separately to the image and LiDAR modalities.

For the image branch, we divide the input into patches and detect boundary patches using a Canny~\cite{canny_edge} edge detector. We fill the drop-out image region with the value 0. 
These edge patches are randomly dropped with a ratio $r^\text{2D}_\text{bd}$.  
For the LiDAR branch, we identify geometric discontinuities by checking label inconsistency between adjacent voxels. 
Voxel features within these discontinuous regions are dropped with a ratio $r^\text{3D}_\text{bd}$, while their coordinates are retained to preserve spatial structure.  
Designing boundary detection separately for the two modalities exposes the model to a broader range of boundary degradation patterns, thereby enriching the diversity of boundary conditions encountered during training. 

\noindent\textbf{Instance Dropout.}  
Maintaining semantic consistency within an instance is crucial for robust perception of thing classes. 
To enhance instance completeness, we perform partial dropout within the interiors of thing-class instances in both modalities.  
In the image branch, a patch inside each instance mask is randomly masked out with ratio $r^\text{2D}_\text{ins}$.
In the LiDAR branch, we randomly select a cluster of points belonging to a thing instance that project onto the same image patch, and zero out their point features with ratio $r^\text{2D}_\text{ins}$, while preserving their coordinates.
This strategy encourages the model to recover complete object representations even when only partial information is available, thereby improving instance-level robustness under domain shifts. 

For each frame, we randomly select one modality (LiDAR or image) with probability $p=0.5$ to apply both instance and boundary dropout strategies.
The dropout ratios are set as $r^\text{2D}_\text{bd}=r^\text{3D}_\text{ins}=r^\text{2D}_\text{ins}=0.5$ and $r^\text{3D}_\text{bd}=0.7$. The patch size is set to $32\times32$.

\noindent\textbf{Auxiliary Loss and Source-only Strategy.}
We apply AMD only on the source domain, aiming to establish an \textit{easy-to-hard} curriculum: the model first learns to handle controlled, synthetic degradations with GT supervision on source data before encountering real-world noise in the target domain, thereby facilitating more stable cross-domain knowledge transfer. 
Notably, PanDA does not require any domain-specific augmentations. The same AMD strategy, with identical hyperparameters, is applied universally across all domain shifts, serving as a simple but effective UDA augmentation. 

To explicitly mask modeling, we attach auxiliary semantic segmentation heads to both the LiDAR and image encoders in the student model. 
The image branch employs an FCN-style head to predict 2D semantic maps, supervised by projected 3D ground-truth labels. The 3D semantic head is implemented via the semantic query module inherently integrated in the base model~\cite{pan_2025_ial}. 
The auxiliary semantic loss $\mathcal{L}_{\text{aux}}^{\mathcal{S}}$ is defined as a cross-entropy objective:
\begin{equation}
\mathcal{L}_{\text{aux}}^{\mathcal{S}} = \mathcal{L}_{\text{ce}}(\tilde{\mathbf{y}}^{\text{2D}}_{\mathcal{S}}, \Pi(\mathbf{y}_\mathcal{S})) + \mathcal{L}_{\text{ce}}(\tilde{\mathbf{y}}^{\text{3D}}_{\mathcal{S}}, \mathbf{y}_\mathcal{S}),
\end{equation}
where $\tilde{\mathbf{y}}^{\text{2D}}_{\mathcal{S}}$ and $\tilde{\mathbf{y}}^{\text{3D}}_{\mathcal{S}}$ denote the predictions from the auxiliary heads, $\Pi$ denotes 3D-to-2D projection. 



\subsection{Pseudo-Label Initiation}
\label{sec:drop}
Existing semantic UDA methods generate pseudo-labels by retaining only high-confidence regions, which breaks object completeness and blurs the mask boundary~\cite{jaritz_xMUDA_2020,wu_2024_unidseg,cardace_mm2d3d_2023,cao_MoPA_2024,jaritz_xMUDAExt_2022}, which severely harms the learning thing and stuff classes. 
To adapt this strategy for panoptic segmentation while minimizing structural damage, we introduce a class-aware filtering approach: For stuff classes, which are uncountable continuous regions like roads or vegetation, we apply region-wise filtering that preserves structural integrity while suppressing local noise. For thing classes, which are compact countable objects, we employ instance-level filtering that removes entire low-confidence instances to avoid false-positive propagation.

The model outputs consist of semantic map $\mathbf{C} \in \mathbb{R}^{K \times C}$ with semantic confidences $\mathbf{S}_C \in \mathbb{R}^{K}$ and instance mask set $\mathbf{M} \in \mathbb{R}^{K \times N}$ with instance confidences $\mathbf{S}_M \in \mathbb{R}^{N}$.  
Here, $C$, $N$, and $K$ denote the number of semantic classes, LiDAR points, and panoptic objects (including all thing classes \{$\mathcal{C}_{\text{thing}}\}$ and stuff classes $\{\mathcal{C}_{\text{thing}}\}$), respectively.  
We compute a joint confidence map $\mathbf{S} \in \mathbb{R}^{N}$ by multiplying each point’s semantic and instance confidence scores, capturing both semantic and instance reliability. 
For \textit{thing} classes, we compute the average confidence per instance: 
\begin{equation}
\vspace{-0.3em}
 \bar{ \mathbf{S}}(k)=\frac{1}{|\mathcal{P}_k|}\sum_{p \in \mathcal{P}_k} \mathbf{S}(p),
\vspace{-0.3em}
\end{equation}
where $\mathcal{P}_k=\{p \mid \hat{\mathbf{M}}_k(p)=1\}$, and remove the entire instance if it is unreliable: 
\begin{equation}
\label{eq:thing_drop}
\mathbf{M}_k^{'}(p) = 
\begin{cases}
1, & \text{if } \mathbf{M}_k(p) = 1 \text{ and }  \bar{ \mathbf{S}}(k) \geq \tau_{th} \\
0, & \text{otherwise,}
\end{cases}
\end{equation}
with $\tau_{th}=0.63$ consistently across all domain shifts. 
For each \textit{stuff} mask $\mathbf{M}_k$, we perform point-wise filtering:
\begin{equation}
\mathbf{M}_k^{'}(p) = 
\begin{cases}
1, & \text{if } \mathbf{M}_k(p) = 1 \text{ and } \mathbf{S}(p) \geq \tau_{st} \\
0, & \text{otherwise,}
\end{cases}
\end{equation} 
where $\tau_{st}$ is adaptively determined per class following~\cite{jaritz_xMUDA_2020,jaritz_xMUDAExt_2022}. 
Points not covered by any mask after filtering are excluded from training by setting their loss weight to zero, preventing erroneous gradient propagation. 

While this initialization reduces noise, three \textbf{limitations} persist: 
a) The stuff mask is incomplete, with hole and broken boundary; 
b) The lack of clear boundary harms the recognition of both thing and stuff class; 
c) The remaining instance mask may be misclassified.

\subsection{\textbf{Dual-Expert Pseudo-Label Refinement} }
\label{sec:PL}
To address the limitations of partial supervision, we propose Dual-Expert Pseudo-Label Refinement (DualRefine), which leverages two complementary domain-invariant experts, 3D geometric and 2D visual priors, to restore instance completeness and correct cross-domain semantic inconsistencies. 
DualRefine consists of two stages: Grow and Reassign, as illustrated in Fig.~\ref{fig:fw}. 
Specifically, we introduce two forms of superpoints as \textit{domain-invariant experts}: 
The \textit{geometric superpoints}, denoted as $\mathcal{G} = \{ g_m \}_{m=1}^{M}$, are extracted directly from LiDAR data and reflect shape-consistent 3D regions and remain stable under severe domain shifts. We use RANSAC to separate ground and non-ground regions, then use HDBSCAN clustering of non-ground points, following the common practice~\cite{largeAD_25Arxiv}.
The \textit{visual superpoints}, denoted as $\mathcal{Q} = \{ q_n \}_{n=1}^{N}$, are obtained by lifting 2D VFM segmentation masks, which bring semantically robust cues from the VFMs. For simplicity, we directly use the point-wise 2D proposals inside IAL, which are predicted by Grounding DINO~\cite{gdino_22eccv} and SAM~\cite{sam_23iccv}, and reprojected to 3D space. 

\noindent\textbf{Grow.} 
To recover the continuity of stuff regions that are broken after confidence-based filtering, we expand each truncated stuff mask $\mathbf{M}_k$ ($k \in \mathcal{C}_{\text{stuff}}$) using the best-matched geometric superpoint $g^*$ that satisfies the overlap constraint:
\begin{equation}
    g^{\star} = \arg\max_{g_m \in \mathcal{G}} \mathrm{IoU}(g_m, \mathbf{M}_k^{'}) \quad \text{s.t.} \quad \mathrm{IoU}(g^{\star}, \mathbf{M}_k^{'}) \geq 0.5.
\end{equation}
The refined stuff mask is obtained by merging two masks:
\begin{equation}
\hat{\mathbf{M}}_k = \mathbf{M}_k^{'} \cup g^*
\end{equation}
When conflicts occur between refined stuff masks $\hat{\mathbf{M}}_k$ and the thing masks, we keep the stuff mask while removing conflicting regions from the thing masks. 

\noindent\textbf{Class Reassignment.} 
To address semantic misclassification caused by cross-domain appearance variations,  we leverage visual superpoints to reassign category labels for low-confidence thing instances. 
Specifically, for each thing instance $k \in \mathcal{C}_{\text{thing}}$, we first find the matched visual superpoint with the same IoU constraint:
\begin{equation}
    q^{\star} = \arg\max_{q_n \in \mathcal{Q}} \mathrm{IoU}(q_n,\hat{\mathbf{M}}_k) \quad \text{s.t.} \quad \mathrm{IoU}(q^{\star}, \hat{\mathbf{M}}_k) \geq 0.5
\end{equation}
Then we use VFM prediction to reassign the model predicted semantic label if the instance confidence is low:
\begin{equation}
\label{eq:reassign}
\hat{\mathbf{C}}_k=
\begin{cases}
\mathbf{C}_{\mathcal{Q}}(q^{\star}), &\text{if } \bar{\mathbf{S}}(k) < \text{min}(\mathbf{S}_{\mathcal{Q}}(q^{\star}) ,t_{\text{cls}}),\\[4pt]
\mathbf{C}_k, & \text{otherwise},
\end{cases}
\end{equation}
where $\mathbf{C}_{\mathcal{Q}}(q^*)$ and $\mathbf{S}_{\mathcal{Q}}(q^*)$ denote the semantic label and confidence from Grounding DINO. $\bar{ \mathbf{S}}(k)$ is the each instance's averaged score and is normalized after filtering in the previous stage.  
We set the selection upper bound as $t_{\text{cls}} = 0.2$.
The final output consists of refined semantic map $\hat{\mathbf{C}}$ and instance mask $\hat{\mathbf{M}}$.

\noindent\textbf{Pseudo-label Refinement Scheme.} 
For each target sample, we first obtain predictions from the teacher model, apply confidence-based filtering in Sec.~\ref{sec:drop} to remove unreliable regions, then refine both masks and labels via the DualRefine procedure. The resulting high-quality pseudo-labels are used to supervise the student model on the target domain, jointly trained with labeled source data. 

%% file: sec/4-experiment.tex
\section{Experimental Results}
\subsection{Experiment Setup}

\begin{table*}[!]
\caption{Comparative and ablative studies for cross-domain 3D multimodal panoptic segmentation.
``Pano-xMUDA'' and ``Pano-UniDseg'' represent adapting 3D semantic UDA work xMUDA~\cite{jaritz_xMUDA_2020} and UniDSeg~\cite{wu_2024_unidseg} to panoptic segmentation. Our method is ablated by DualRefine, AMD modules or both (-base). Top and runner‑up results are marked in \textbf{bold} and \underline{underline}, respectively.}
\label{tab:main}
  \vspace{-0.1in}
\renewcommand\tabcolsep{3.0pt}
\resizebox{\linewidth}{!}{
\begin{tabular}{l cccc cccc cccc cccc}
\toprule[1.2pt]
\multirow{2}{*}{\textbf{Method}} & \multicolumn{4}{c}{\textbf{nuSc.: USA/SG}} & \multicolumn{4}{c}{\textbf{nuSc.: Sunny/Rainy}} & \multicolumn{4}{c}{\textbf{nuSc.: Day/Night}} & \multicolumn{4}{c}{\textbf{Sem.KITTI/nuSc.}} \\ 
\cmidrule(lr){2-5} \cmidrule(lr){6-9} \cmidrule(lr){10-13} \cmidrule(lr){14-17}
& PQ & mIoU & $\text{PQ}^\text{th}$ & $\text{PQ}^\text{st}$ & PQ & mIoU & $\text{PQ}^\text{th}$ & $\text{PQ}^\text{st}$ & PQ & mIoU & $\text{PQ}^\text{th}$ & $\text{PQ}^\text{st}$ & PQ & mIoU & $\text{PQ}^\text{th}$ & $\text{PQ}^\text{st}$ \\ 
\midrule
Baseline                                     & 64.1   & 58.3  & 64.4    & 63.5    & 63.5    & 59.7   & 65.9     & 62.1     & \multicolumn{1}{c}{64.7} & \multicolumn{1}{c}{56.1} & \multicolumn{1}{c}{58.0} & \multicolumn{1}{c|}{75.9} & 1.2   & 7.6   & 1.8     & 1.1     \\
Pano-xMUDA                                    & 67.2          & 63.4          & 68.6          & 64.9          & 62.2          & 57.3          & 55.8          & \underline{72.8}          & 69.7                     & \underline{60.3}                     & 74.2                     & 64.4                      & 49.1          & 59.3          & 59.8          & 47.0          \\
Pano-UniDSeg                                  & 72.9          & 68.1          & 71.3          & \textbf{75.4} & 65.5          & 57.6          & 60.6          & \textbf{73.7} & 70.5                     & \textbf{61.0}            & 75.3                     & \underline{64.9}                      & \underline{54.0}          & 61.9          & 58.0          & \textbf{53.2} \\\hline
Ours-base                                     & 73.7          & 70.3          & 74.9          & 71.9          & 70.3          & 65.7          & 69.9          & 71.5          & 68.6                     & 58.9                     & 72.2                     & 64.5                      & 51.6          & 63.9          & 61.1          & 49.7          \\
Ours-DualRefine                               & 75.6          & \underline{71.6}          & 77.2          & 73.1          & 71.6          & 66.5          & \underline{71.7}          & 71.3          & \underline{71.2}                     & 59.0                     & \underline{76.8}                     & 64.6                      & \underline{54.0}          & \underline{66.1}          & \underline{68.4}          & 51.1          \\
Ours-AMD                                      & \underline{76.8}          & 71.5          & \underline{78.7}          & 73.9          & \underline{71.8}          & \underline{67.6}          & 71.4          & 72.4          & 70.9                     & 57.4                     & 76.6                     & 64.2                      & 53.5          & 65.5          & 67.8          & 50.6          \\
\rowcolor{gray!10}Ours-Final                                   & \textbf{77.3}   & \textbf{72.3}  & \textbf{79.3}    & \underline{74.2}    & \textbf{72.4}    & \textbf{67.9}   & \textbf{72.3}     & {72.4}     & \textbf{73.1}                     & 59.9                     & \textbf{79.6}                     & \textbf{65.5}                      & \textbf{54.5}  & \textbf{66.4}  & \textbf{70.4}    & \underline{51.3}    \\\hline
Oracle-Target                         & 79.9   & 77.7  & 80.4    & 78.9    & 74.4    & 67.8   & 75.5     & 72.5     & 53.5                     & 51.9                     & 49.1                     & 58.8                      & 81.5  & 86.1  & 93.7    & 79.0    \\
Oracle-Joint                    & 82.0   & 77.8  & 83.5    & 79.4    & 81.2    & 76.4   & 85.4     & 74.3     & 70.6                     & 60.4                     & 76.8                     & 63.3                      & 81.8  & 86.8  & 93.9    & 79.4    \\ \bottomrule[1.2pt]
\end{tabular}
}
\end{table*}

\textbf{Intra-Dataset Domain Shifts.}
We evaluate three types of intra-dataset domain shifts derived from the nuScenes dataset~\cite{nuscenes,nuscenes_panoptic}, which provides data from a 32-beam LiDAR and six RGB cameras. 
Based on the official split, we divide nuScenes into domains according to metadata attributes (Day \vs Night, Sunny \vs Rainy, and Boston \vs Singapore), to examine realistic variations in illumination, weather, and location. 
Following standard multi-modal panoptic segmentation settings~\cite{pan_2025_ial,lcps_23iccv,p.-fusionnet_24ESA}, we use LiDAR points across the full range and all six camera views, and remap all categories into 10 ``thing'' and 6 ``stuff'' classes. 

\noindent\textbf{Cross-Dataset Domain Shift.}
To further examine cross-dataset performance, we conduct experiments from SemanticKITTI~\cite{skitti_behley2019iccv,skitti_behley2021ijrr} to nuScenes. 
SemanticKITTI is collected in Germany using a 64-beam LiDAR and two front-view cameras, which introduces larger discrepancies in both sensors and scene characteristics compared to nuScenes. 
We follow~\cite{jaritz_xMUDAExt_2022} for class mapping, resulting in 1 ``thing'' and 5 ``stuff'' categories.
Detailed label mappings and dataset statistics are provided in the supplementary material. 

\noindent\textbf{Implementation Details. } 
We follow the standard panoptic segmentation pipeline~\cite{pan_2025_ial,p-polarnet_21cvpr,p3former_25ijcv} for data preprocessing, where LiDAR point clouds are discretized into cylindrical voxels of size  $[480 \times 360 \times 32]$ and images at all views are resized into $640 \times 360$. 
We initialize both the teacher and student networks using the panoptic segmentation model pretrained on the source domain using the same augmentation and training scheme as IAL~\cite{pan_2025_ial}. 
During UDA training, source data are supervised by GT labels, while target data are supervised using pseudo-labels refined by DualRefine strategy. 
In addition to the proposed AMD augmentation, we also apply PieAug~\cite{pan_2025_ial} within each domain to enhance data diversity. 
To handle varying dataset sizes, the total training iterations are set to $D \times \text{epoch\_len}$, where $\text{epoch\_len}$ equals the number of training samples in the target domain. 
We set $D=30$ for Day→Night and Sunny→Rainy, and $D=15$ for USA→Singapore and SemanticKITTI→nuScenes. 
Training follows an iteration-based schedule with an initial learning rate of 0.0004, reduced by half at 1/3, 1/2, and 2/3 of the total iterations.
We use the AdamW optimizer~\cite{adamw} with a weight decay of 0.01, and update the teacher model at every iteration via EMA with a momentum of 0.99.
All experiments are conducted with a batch size of 2 on four NVIDIA A40 or H100 GPUs.

\noindent\textbf{Evaluation Metrics.} 
Consistent with the standard panoptic segmentation,  we use panoptic quality (PQ)  as  the primary metric~\cite{PQ}. PQ is defined as the product of segmentation quality (SQ) and recognition quality (RQ):
\begin{equation}
    \text{PQ} = \underbrace{\frac{\sum_{\text{TP}} \text{IoU}}{|\text{TP}|}}_{\text{SQ}} \times 
    \underbrace{\frac{|\text{TP}|}{|\text{TP}| + \frac{1}{2} |\text{FP}| + \frac{1}{2} |\text{FN}|}}_{\text{RQ}},
\end{equation}
These metrics can be further extended to ``thing'' and ``stuff'' classes, denoted as 
$\text{PQ}^{\text{th}}$ and $\text{PQ}^{\text{st}}$.

\subsection{Baselines}
To demonstrate the effectiveness of our method, we include both Baselines and Oracles for comparison.
The \textit{\underline{Baseline}} is trained on labeled source data and evaluated on the target domain, serving as a lower bound that reflects the domain gap.
\textit{\underline{Oracle-Target}} (trained and evaluated on the target domain) provides an upper bound, while \textit{\underline{Oracle-Joint}} (trained on both source and target data) represents supervised cross-domain performance. 
Notably, the Oracle-Target for Day→Night result (53.5\%) is lower than the Baseline (64.7\%) due to the limited size of the night domain (602 frames), though Oracle-Joint remains a reliable reference. 
As the first work on multimodal 3D panoptic UDA, we adapt two semantic UDA methods, xMUDA~\cite{jaritz_xMUDA_2020} and UniDSeg~\cite{wu_2024_unidseg}, into our experimental setting. 
Specifically, \textit{\underline{Pano-xMUDA}} 
utilizes the IAL \cite{pan_2025_ial} backbone while incorporating an additional 2D semantic head and two mimic heads (2D and 3D) to compute cross-modal loss alongside the native 3D semantic output.
\textit{\underline{Pano-UniDSeg}} integrates the CLIP~\cite{clip} ViT-L backbone for the 2D branch and applies their proposed Modal Transitional Prompting and Learnable Spatial Tunability modules for multimodal interaction, using the cross-modal loss and the panoptic backbone IAL same as Pano-xMUDA. We also provide component results \textit{\underline{Ours-DualRefine}} and \textit{\underline{Ours-AMD}}  for comparisons. And \textit{\underline{Ours-base}} denotes using only the  mean-teacher framework without any proposed designs. All component results are trained under the same conditions.  

\subsection{Main Results}
We report comprehensive results under various domain-shift scenarios in Table~\ref{tab:main}.
Compared the final framework (\textit{\underline{Ours-Final}}, colored in gray) with Baseline (at row 1), the synergy of our proposed modules DualRefine and AMD achieves consistent and substantial improvements across all domain shifts, outperforming the corresponding Baseline by +13.2\%, +8.9\%, +8.4\% and +53.3\% on PQ for the USA→SG, Sunny→Rainy,  Day→Night, and SemanticKITTI→nuScenes settings, respectively. These results demonstrate the strong robustness and generalization capability of our method under diverse domain shifts. 
Remarkably, our model surpasses the oracle upper-bound (row 8 and 9) in the Day→Night setting, confirming that the proposed augmentation and pseudo-label refinement modules together bridge the supervision gap under severe conditions. 

Our method consistently outperforms both adapted semantic UDA approaches Pano-xMUDA and Pano-UniDSeg across all domain shifts, with particularly substantial gains on thing classes, underscoring the effectiveness of our design in instance-level perception. These gains are attributed to our two core components: AMD enhances domain-invariant feature learning from the source domain, while DualRefine recovers reliable instance structures in the target domain. Notably, performance improvements are most pronounced under challenging shifts the Day→Night and Sunny→Rainy, validating our framework's robustness to severe modality degradation.

\begin{figure*}[htb]
    \centering
    \includegraphics[width=\linewidth]{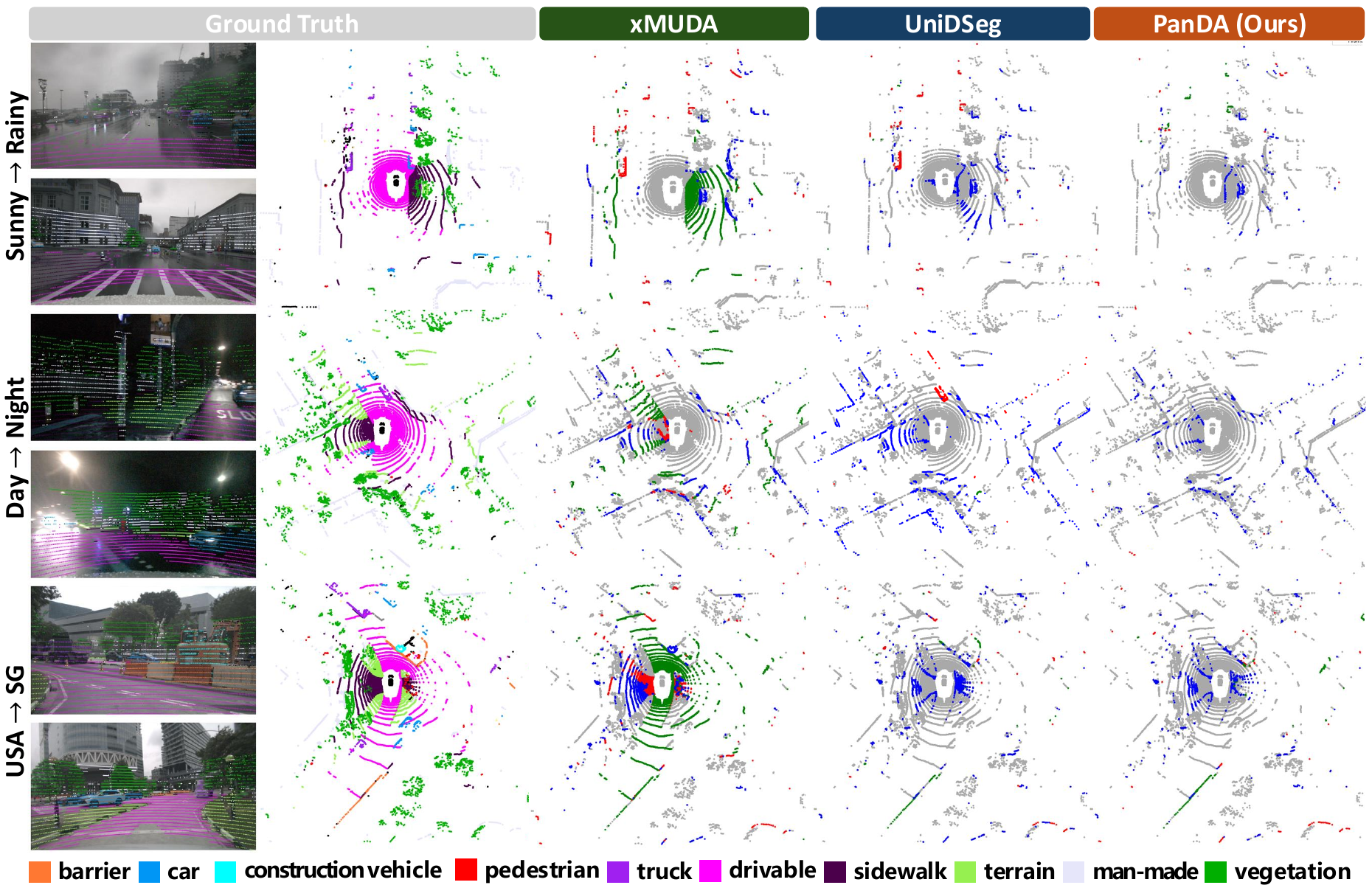}
    \vspace{-0.2in}
    \caption{
Visualization of panoptic segmentation error maps across various domain shifts, comparing our method with xMUDA and UniDSeg. 
Point colors indicate error types: \textcolor{red}{\textbf{false positives}} and \textcolor[rgb]{0.21,0.463,0.13}{\textbf{false negatives}} reflect recognition errors, while \textcolor[rgb]{0.663,0.663,0.663}{\textbf{well-matched}} and \textcolor{blue}{\textbf{mismatched}} points (within true positives) highlight segmentation inaccuracies. 
Raw point clouds, camera images, and semantic-colored ground truth (GT) are shown for reference. Best viewed in color.
}
    \label{fig:vis}
\end{figure*}

Fig. \ref{fig:vis} presents panoptic error maps across multiple domain shifts. 
These visual comparisons confirm that PanDA significantly reduces recognition errors (false positives/negatives) and segmentation errors (mismatched areas), validating its strong cross-domain generalization capability.

\begin{table}[!]
\small
\caption{Ablation study of the AMD module. “Ins.” and “Bd.” indicate instance and boundary drop, and “Loss” denotes the use of auxiliary semantic loss.}
  \vspace{-0.1in}
\label{tab:ablt-asym}
\setlength{\tabcolsep}{3pt}
\resizebox{\linewidth}{!}{
\begin{tabularx}{\columnwidth}{p{0.03\columnwidth}p{0.03\columnwidth}p{0.05\columnwidth}YYY YYY YYY}
\toprule[1.2pt]

\multicolumn{2}{c}{Aug} &
\multicolumn{1}{c}{\multirow{2}{*}{Loss}} &
\multicolumn{3}{c}{\textbf{USA/SG}} &
\multicolumn{3}{c}{\textbf{Sunny/Rainy}} &
\multicolumn{3}{c}{\textbf{Day/Night}} \\

\cmidrule(lr){1-2}
\cmidrule(lr){4-6}
\cmidrule(lr){7-9}
\cmidrule(lr){10-12}

Ins. & Bd. & 
& PQ & $\text{PQ}^{\text{th}}$ & $\text{PQ}^{\text{st}}$
& PQ & $\text{PQ}^{\text{th}}$ & $\text{PQ}^{\text{st}}$
& PQ & $\text{PQ}^{\text{th}}$ & $\text{PQ}^{\text{st}}$ \\

\midrule
            &           &                                            & 75.6     & 77.2       & 73.1       & 71.6       & 71.7        & 71.3        & 71.2      & 76.8        & 64.6       \\
            &           & \ding{51}                                          & 76.9     & 78.7       & 74.3       & 71.7       & 71.3        & 72.3        & 67.9      & 70.0        & 65.5       \\
\ding{51}           &           & \ding{51}                                          & 77.0     & 79.0       & 74.1       & 72.0       & 71.6        & 72.6        & 69.4      & 72.9        & 65.3       \\
            & \ding{51}         & \ding{51}                                          & 76.8     & 78.6       & 74.1       & 72.3       & 72.2        & 72.5        & 70.8      & 75.2        & 65.7       \\
\ding{51}           & \ding{51} & & 77.1     & 79.1       & 74.1       & 71.9       & 71.8        & 72.4        & 71.6      & 77.9        & 64.2       \\ 

\ding{51}           & \ding{51}         & \ding{51}                                          & 77.3     & 79.3       & 74.2       & 72.4       & 72.3        & 72.4        & 73.1      & 79.6        & 65.5       \\ \bottomrule[1.2pt]
\end{tabularx}}
\end{table}

\begin{table}[!]
\caption{Ablation of the pseudo-label strategies. The drop denotes the confidence-based filtering for pseudo-label. The grow and class denote two components in our proposed DualRefine.}
  \vspace{-0.1in}
\label{tab:ablt-pl}
\small
\setlength{\tabcolsep}{3pt}
\begin{tabularx}{\columnwidth}{p{0.015\columnwidth}>{\centering\arraybackslash}p{0.015\columnwidth}>{\centering\arraybackslash}p{0.015\columnwidth}YYYYYY}
\toprule[1.2pt]
\multicolumn{3}{c}{DualRefine}                                                      & \multicolumn{2}{c}{\textbf{USA/SG}}                        & \multicolumn{2}{c}{\textbf{Sunny/Rainy}}                    & \multicolumn{2}{c}{\textbf{Day/Night}}                     \\ 
\cmidrule(lr){1-3}
\cmidrule(lr){4-5}
\cmidrule(lr){6-7}
\cmidrule(lr){8-9}

\multicolumn{1}{c}{drop} & \multicolumn{1}{c}{grow} & \multicolumn{1}{c}{class} & \multicolumn{1}{c}{PQ} & \multicolumn{1}{c}{mIoU} & \multicolumn{1}{c}{PQ} & \multicolumn{1}{c}{mIoU} & \multicolumn{1}{c}{PQ} & \multicolumn{1}{c}{mIoU} \\ \midrule
                         &                          &                               & 73.7                   & 70.3                      & 70.3                   & 65.7                      & 68.6                   & 58.9                     \\
\ding{51}                        &                          &                               & 74.8                   & 71.2                      & 70.0                   & 66.0                      & 66.5                   & 58.7                     \\
                         & \ding{51}                        &                               & 75.3                   & 70.6                      & 70.9                   & 66.2                      & 70.8                   & 59.7                     \\
                         &                          & \ding{51}                             & 75.4                   & 71.5                      & 71.1                   & 66.5                     & 69.9                   & 58.3                     \\
\ding{51}                        & \ding{51}                        & \ding{51}                             & 75.6                   & 71.6                      & 71.6                   & 66.5                      & 71.2                   & 59.0                     \\ \bottomrule[1.2pt]
\end{tabularx}
\vspace{-1em}
\end{table}

\subsection{Ablation Studies}

To comprehensively evaluate the effectiveness of our proposed components AMD and DualRefine, we conduct detailed ablation studies across three distinct domain shifts, with results summarized in  Tables~\ref{tab:ablt-asym} and \ref{tab:ablt-pl}. All experiments employ identical hyperparameter settings to ensure fair comparisons. 
Table \ref{tab:ablt-asym} provides a comprehensive analysis of AMD. Employing the auxiliary semantic losses improves ``stuff'' category perception (+1.2\%, +1.0\%, and +0.9\% $\text{PQ}^\text{st}$ across the three domain shifts, respectively), but causes performance degradation in ``thing'' classes under challenging conditions—most notably a 6.8\% drop in the Day/Night shift. This suggests that additional supervision alone is insufficient when a specific modality undergoes significant degradation. 
The introduction of instance and boundary drop (Rows 3, 4, and 6) effectively rectifies this limitation, yielding substantial gains in ``thing'' class performance. Specifically, Row 6 achieves an improvement of up to +9.6\% $\text{PQ}^\text{th}$ in the Day/Night shift compared to Row 2. Furthermore, the AMD-only results in Row 5 show significant improvements in "thing" classes that offset the decline observed in the semantic loss-only baseline (Row 2). This underscores the necessity of combining these elements to achieve consistent improvements across all categories. 

\begin{figure}[t]
    \centering
    \vspace{-5mm}
    \includegraphics[width=\linewidth]{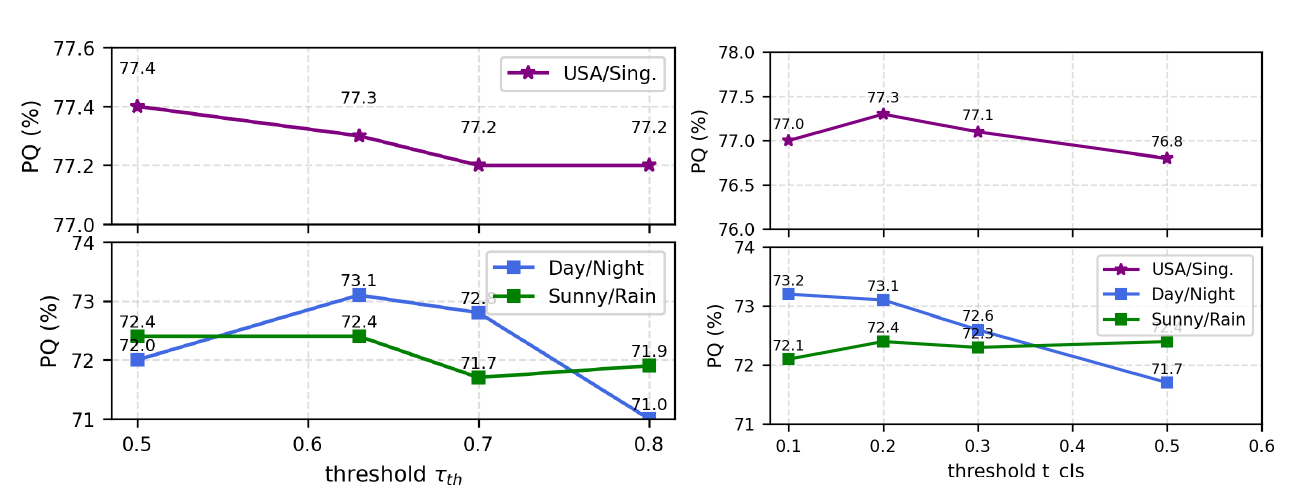}
      \vspace{-2em}
    \caption{Performance variation with drop ratio $\tau_{\text{th}}$ in pseudo-label initiation and class reassignment threshold $t_{cls}$ in DualRefine.}
    \label{fig:t-cls}
    \vspace{-2em}
\end{figure}
The results in Table \ref{tab:ablt-pl} validate the effectiveness of our DualRefine module. Conventional thresholding leads to performance degradation on both Sunny/Rainy (-0.3\%) and Day/Night (-2.1\%) shifts, highlighting the inherent limitation of partial supervision. Applying either the Grow or Class Reassignment stage individually improves performance across all three domain shifts, while combining both yields the highest gains (+1.9\%, +1.3\% and +2.6\%), demonstrating the synergistic effect of integrating 2D and 3D priors for pseudo-label refinement.

We further analyze the sensitivity of the hyperparameters $\tau_{\text{th}}$ and $t_{cls}$ defined in Eq.~\ref{eq:thing_drop} and Eq.~\ref{eq:reassign}, respectively. As shown in Fig.~\ref{fig:t-cls}, our method maintains stable performance across a wide range of values under various domain shifts, demonstrating its robustness to these hyperparameters. 

\section{Conclusion}


We presented PanDA, the first framework to address unsupervised domain adaptation for multimodal 3D panoptic segmentation. To tackle domain discrepancies and unreliable target annotations, PanDA introduces two key components: asymmetric multimodal drop, an augmentation strategy that simulates modality degradation in the source domain to promote domain-invariant representation learning; and DualRefine, a pseudo-label refinement module that exploits complementary 2D visual and 3D geometric priors to produce more complete and reliable supervision in the target domain. Extensive experiments across diverse domain shifts demonstrate PanDA’s effectiveness and strong cross-domain robustness for 3D panoptic perception.
\newpage